\documentclass{llncs}
\usepackage[utf8]{inputenc}
\usepackage{amsmath}
\usepackage{booktabs}
\begin{document}

\mainmatter

\title{Identification of Dynamical Systems using Symbolic Regression}

\author{Gabriel~Kronberger\orcidID{0000-0002-3012-3189} \and Lukas~Kammerer\orcidID{0000-0001-8236-4294} \and Michael~Kommenda}
\authorrunning{Gabriel Kronberger et al.}

\institute {
    Josef Ressel Centre for Symbolic Regression\\
    Heuristic and Evolutionary Algorithms Laboratory\\
    University of Applied Sciences Upper Austria, Softwarepark 11, 4232 Hagenberg\\
    \email{gabriel.kronberger@fh-hagenberg.at}
}

\maketitle

\begin{abstract}
  We describe a method for the identification of models for dynamical
  systems from observational data. The method is based on the concept
  of symbolic regression and uses genetic programming to evolve a
  system of ordinary differential equations (ODE).

  The novelty is that we add a step of gradient-based optimization of
  the ODE parameters. For this we calculate the sensitivities of the
  solution to the initial value problem (IVP) using automatic
  differentiation.

  The proposed approach is tested on a set of 19 problem instances
  taken from the literature which includes datasets from simulated
  systems as well as datasets captured from mechanical systems. We
  find that gradient-based optimization of parameters improves
  predictive accuracy of the models. The best results are obtained
  when we first fit the individual equations to the numeric
  differences and then subsequently fine-tune the identified parameter
  values by fitting the IVP solution to the observed variable values.
\end{abstract}


\section{Background and Motivation}

Modelling, analysis and control of dynamical systems are core topics
within the field of system theory focusing on the
behavior of systems over time. System dynamics can be modelled using
ordinary differential equations (cf. \cite{Isermann2011}) which define
how the state of a system changes based on the current state for
infinitesimal time steps.


Genetic programming (GP) is a specific type of evolutionary
computation in which computer programs are evolved to solve a given
problem. Symbolic regression (SR) is a specific task for which GP has
proofed to work well. The goal in SR is to find an expression that
describes the functional dependency between a dependent variable and
multiple independent variables given a dataset of observed values for
all variables. In contrast to other forms of regression analysis it is
not necessary to specify the model structure beforehand because the
appropriate model structure is identified simultaneously with the
numeric parameters of the model. SR is therefore especially suited for
regression tasks when a parametric model for the system or process is
not known.
Correspondingly, SR could potentially be used for the identification
models for dynamical systems when an accurate mathematical model of
the system doesn't exist.


We aim to use SR to find the right-hand sides $f(\cdot)$ of a system
of ordinary differential equations such as:
\begin{equation*}
\dot{u_1} = f_{\dot{u_1}}(u_1, u_2, \vec{\theta}), \dot{u_2} = f_{\dot{u_2}}(u_1, u_2, \vec{\theta})
\end{equation*}


We only consider systems without input variables
(non-forced systems) and leave an analysis for forced systems for
future work.


\subsection{Related Work}
The vast literature on symbolic regression is mainly focused on static
models in which predicted values are independent given the input
variable values. However, there are a several articles which
explicitly describe GP-based methods for modelling dynamical
systems. A straight-forward approach which can be implemented
efficiently is to approximate the derivatives numerically
\cite{Iba2008,Schmidt2008,Gaucel2014,worm2014}.  Solving the IVP for
each of the considered model candidates is more accurate but also
computationally more expensive; this is for instance used in
\cite{bongard2007}.

In almost all methods discussed in prior work the individual equations
of the system are encoded as separate trees and the SR solutions
hold multiple trees \cite{Iba2008,Schmidt2008,bongard2007}. A notable
exception is the approach suggested in \cite{Gaucel2014} in which GP
is run multiple times to produce multiple expressions for the
numerically approximated derivatives of each state
variable. Well-fitting expressions are added to a pool of
equations. In the end, the elements from the pool are combined and the
best-fitting ODE system is returned. In this work the IVP is solved
only in the model combination phase.

Identification of correct parameter values is critical especially for
ODE systems. Therefore, several authors have included gradient-based
numeric optimization of parameter values. In \cite{Iba2008},
parameters are allowed only for scaling top-level terms, to allow
efficient least-squares optimization of parameter values. In
\cite{worm2014} non-linear parameters are allowed and parameter values
are iteratively optimized using the Levenberg-Marquardt algorithm
based on gradients determined through automatic differentiation.

Partitioning is introduced in \cite{bongard2007}. The idea is to
optimize the individual parts of the ODE system for each state
variable separately, whereby it is assumed that the values of all
other state variables are known. Partitioning reduces computational
effort but does not guarantee that the combined system of equations
models system dynamics correctly.


Recently, neural networks have been used for modelling ODEs
\cite{Chen2018} whereby the neural network parameters are optimized
using gradients determined via the adjoint sensitivity method and
automatic differentiation. The same idea can be used to fit numeric
parameters of SR models.

\section{Methods}
We extend the approach described in \cite{Iba2008}, whereby we allow
numeric parameters at any point in the symbolic expressions. We use an
iterative gradient-based method (i.e. Levenberg-Marquardt (LM)) for
the optimization of parameters and automatic differentiation for
efficient calculation of gradients similarly to \cite{worm2014}. We
analyse several different algorithm variations, some of which solve
the IVP problem in each evaluation step as in
\cite{bongard2007,Chen2018}. For this we use the state-of-the-art
CVODES\footnote{https://computation.llnl.gov/projects/sundials/cvodes}
library for solving ODEs and calculating parameter sensitivities.


We compare algorithms based on the deviation of the IVP solution from the observed data points for
the final models. The deviation measure
is the sum of normalized mean squared errors:
\begin{equation}
  \operatorname{SNMSE}(Y, \hat{Y}) = \sum_{i=1}^D \frac{1}{\operatorname{var}(y_{i,\cdot})} \frac{1}{N} \sum_{j = 1}^{N} (y_{i,j} - \hat{y}_{i,j})^2
  \label{eqn:ode-snmse}
\end{equation}
Where $Y=\{y_{i,j}\}_{i=1..D,j=1..N}$ is the matrix of $N$ subsequent
and equidistant observations of the $D$ the state variables. Each
state variable is measured at the same time points
$(t_i)_{i=1..N}$. The matrix of predicted values 
variables $\hat{Y} = \{\hat{y}_{i,j}\}_{i=1..D, j=1..N}$ is calculated
by integrating the ODE system
using the initial values $\hat{y}_{\cdot, 1} = y_{\cdot, 1}$

\subsection{Algorithm description}
The system of differential equations is represented as an array of
expression trees each one representing the differential equation for
one state variable.  The evolutionary algorithm initializes each tree
randomly whereby all of the state variables are allowed to occur in
the expression. In the crossover step, exchange of sub-trees is only
allowed between corresponding trees. This implicitly segregates the
gene pools for the state variables. For each of the trees within an
individual we perform sub-tree crossover with the probability given by
the crossover rate parameter. A low crossover rate is helpful to
reduce the destructive effect of crossover.

Memetic optimization of numeric parameters has been shown to improve
GP performance for SR \cite{topchy2001,kommenda2013}. We found that GP
performance is improved significantly even when we execute only a few
iterations (3-10) of LM and update the parameter values 
whenever an individual is evaluated. This ensures that the improved
parameter values are inherited and can subsequently be improved
even further when we start LM with these values \cite{kommenda2013}.
We propose to use the same approach for parameters of the ODEs with a small modification
based on the idea of partitioning. In the first step we use the approximated derivative
values as the target. In this step we
use partitioning and assume all variable values are known. The parameter values are updated using the optimized values. In the
second step we use CVODES to solve the IVP for all state variables
simultaneously and to calculate parameter sensitivities. The LM algorithm is used to optimize the
SNMSE (Equation \ref{eqn:ode-snmse}) directly. Both steps can
be turned on individually and the number of LM iterations can be set
independently.


\subsection{Computational experiments}

\subsubsection{Algorithm configuration}
Table \ref{tab:algorithm-configuration} shows the GP parameter values
that have been used for the experiments.
\begin{table}
  \centering
  \begin{tabular}{lp{7.3cm}}
    Parameter & Value \\
    \toprule
    Population size & 300 \\
    Initialization & PTC2 \\
    Parent selection & proportional (first parent) \newline random (second parent)\\
    Crossover & Subtree crossover \\
    Mutation & Replace subtree with random branch \newline Add $x\sim N(0, 1)$ to all numeric parameters \newline Add $x\sim N(0, 1)$ to a single numeric parameter.\newline Change a single function symbol.  \\ 
    Crossover rate & 30\% (for each expression)\\
    Mutation rate & 5\% (for the whole individual)\\
    Offspring selection & offspring must be better than both parents \\
    Maximum selection pressure & $100 < $ \# evaluated offspring / population size\\  
    Replacement & Generational with a single elite.\\
    Terminal set & State variables and real-valued parameters\\
    Function set & $+, *, \sin, \cos, $ \\
    \bottomrule
  \end{tabular}
  \caption{\label{tab:algorithm-configuration}Parameter values for the
    GP algorithm that have been used for all experiments. The number
    of generations and the maximum number of evaluated solutions is
    varied for the experiments.}
\end{table}

We compare two groups of different configurations: in the first group
we rely solely on the evolutionary algorithm for the identification of
parameter values. In the second group we use parameter optimization
optimization. Both groups contain three configurations with different
fitness functions. In the following we use the identifiers D, I, D+I,
D$_{\operatorname{opt}}$, I$_{\operatorname{opt}}$, and
D$_{\operatorname{opt}}$+I$_{\operatorname{opt}}$ for the six
algorithm instances. Configuration D uses the SNMSE 
for the approximated derivatives for fitness assignment as in
\cite{Iba2008,Gaucel2014,worm2014}; configuration I uses the SNMSE for
the solution to the IVP as in \cite{bongard2007}; configuration D+I
uses the sum of both error measures for fitness evaluation. For the
first group we allow maximally 500.000 evaluated solutions and 250
generations; for the second group we only allow 100.000 evaluated
solutions and 25 generations. The total number of function evaluations
including evaluations required for parameter optimization is similar
for all configurations and approximately between 500.000 and 2 million
(depending on the dimensionality of the problem).


\subsubsection{Problem instances}
We use 19 problem instances for testing our proposed approach as shown
in Table \ref{tab:synthetic-instances} and Table
\ref{tab:real-instances}. These have been taken from
\cite{Iba2008,Schmidt2008,Schmidt2009} and include a variety of
different systems.  The set of problem instances includes simulated
systems (Table \ref{tab:synthetic-instances}) as well as datasets
gathered with motion-tracking from real mechanical systems (Table
\ref{tab:real-instances}). The simulated datasets have been generated
using fourth-order Runge-Kutta integration (RK45).The motion-tracked
datasets have been adapted from the original source to have
equidistant observations using cubic spline interpolation.

\begin{table}
  \centering
  \begin{tabular}{p{2.5cm}cccc}
    Instance & Expression & Initial value & $N$ & $t_{max}$ \\
    \toprule
     Chemical\newline reaction \cite{Iba2008} &
    \def\arraystretch{2}
    $ \begin{aligned}
      \dot{y_1} & = -1.4 y_1 \\
      \dot{y_2} & = 1.4 y_1 -4.2 y_2 \\
      \dot{y_3} & = 4.2 y_2
    \end{aligned}$
    & $(0.1, 0, 0)$ &  $100$ & $ 1$ \\
    \midrule
    
     E-CELL\newline \cite{Iba2008} &
    \def\arraystretch{2}
    $\begin{aligned}
      \dot{y_1} & = -10 y_1 y_3 \\
      \dot{y_2} & = 10 y_1 y_3 -17 y_2 \\
      \dot{y_3} & = -10 y_1 y_3 + 17 y_2
    \end{aligned}$
    & $(1.2, 0.0, 1.2)$ & $40$ & $0.4$ \\
    \midrule
    
    S-System\newline \cite{Iba2008} &
    $\begin{aligned}
      \dot{y_1} & = 15 y_3 y_5^{-0.1} -10 y_1^2 \\ 
      \dot{y_2} & = 10 y_1^2 -10 y_2^2 \\
      \dot{y_3} & = 10 y_2^{-0.1} -10 y_2^{-0.1} y_3^2\\
      \dot{y_4} & = 8 y_1^2 y_5^{-0.1} -10 y_4^2 \\
      \dot{y_5} & = 10 y_4^2 -10 y_5^2
    \end{aligned}$
    & $\begin{aligned}
      (0.1, 0.1, 0.1, 0.1, 0.1)\\
      (0.5, 0.5, 0.5, 0.5, 0.5)\\
      (1.5, 1.5, 1.5, 1.5, 1.5)
     \end{aligned}$ & $3 * 30$ & $0.3$ \\
     \midrule

     Lotka-Volterra \newline (3 species) \cite{Iba2008} &
    $\begin{aligned}
      \dot{y_1} & = y_1 (1 - y_1 - y_2 -10 y_3) \\
      \dot{y_2} & = y_2 (0.992 -1.5 y_1 - y_2 - y_3) \\
      \dot{y_3} & = y_3 (-1.2 + 5 y_1 + 0.5 y_2) 
    \end{aligned}$
    & $(0.2895, 0.2827, 0.126)$ & $100$ & $100$ \\
    \midrule
    
     Lotka-Volterra \newline (2 species) \cite{Gaucel2014} &
    $\begin{aligned}
      \dot{y_1} & = y_1 (0.04 - 0.0005 y_2) \\
      \dot{y_2} & = - y_2 (0.2 - 0.004 y_1)
    \end{aligned}$
    & $(20, 20)$ & $300$ & $300$ \\
    \midrule

     Glider \newline \cite{Schmidt2008} &
    $\begin{aligned}
      \dot{v} & = -0.05 v^2 - \sin(\theta)\\
      \dot{\theta} & = v - \cos(\theta) / v; 
    \end{aligned}$
    & $(1.5, 1)$ & $100$ & $10 $ \\
    \midrule

     Bacterial \newline respiration \cite{Schmidt2008} &
    $\begin{aligned}
      \dot{x} & = (20 - x - x y) / (1 + 0.5 x^2) \\
      \dot{y} & = (10 - x y ) / ( 1 + 0.5 x^2) 
    \end{aligned}$
    & $(1, 1)$ & $100$ & $10 $ \\
    \midrule

     Predator-Prey \newline  \cite{Schmidt2008} &
    $\begin{aligned}
      \dot{x} & = x (4 - x - y / (1 + x)) \\
      \dot{y} & = y ( x / (1 + x) - 0.075 y)
    \end{aligned}$
    & $(1.1, 7.36)$ & $100$ & $10 $ \\
    \midrule

     Bar magnets  \newline \cite{Schmidt2008} &
    $\begin{aligned}
      \dot{\theta_1} & = 0.5 \sin(\theta_1 - \theta_2) - \sin(\theta_1) \\
      \dot{\theta_2} & = 0.5 \sin(\theta_2 - \theta_1) - \sin(\theta_2)
    \end{aligned}$
    & $(0.7, -0.3)$ & $100$ & $10 $ \\
    \midrule

     Shear flow  \newline \cite{Schmidt2008} &
    $\begin{aligned}
      \dot{\theta} & = \operatorname{cot}(\theta) \cos(\phi) \\
      \dot{\phi} & = (\cos(\phi)^2 + 0.1\sin(\phi)^2) \sin(\phi)
    \end{aligned}$
    & $(0.7, 0.4)$ & $100$ & $10 $ \\
    \midrule

     Van der Pol  \newline Oscillator \cite{Schmidt2008} &
    $\begin{aligned}
      \dot{x} & = 10 (y - (\frac{1}{3}x^3 - x)) \\
      \dot{y} & = -0.1 x
    \end{aligned}$
    & $(2, 0.1)$ & $100$ & $10$ \\
    \bottomrule
  \end{tabular}
  \caption{\label{tab:synthetic-instances}The problem instances for which we have generated data using numeric integration.} 
\end{table}

\begin{table}
  \centering
  \begin{tabular}{lllcc}
    Type & Name & File name & Variables & $N$ \\
    \toprule
    Simulated & Linear oscillator & \texttt{linear\_h\_1.txt} & $x, v$ & $512$ \\
    Motion-tracked & Linear oscillator & \texttt{real\_linear\_h\_1.txt} & $x, v$ & $879$ \\
    Simulated & Pendulum & \texttt{pendulum\_h\_1.txt} & $\theta, \omega $ & $502$ \\
    Motion-tracked & Pendulum & \texttt{real\_pend\_h\_1.txt} & $\theta, \omega$ & $568$ \\
    Simulated & Coupled oscillator & \texttt{double\_linear\_h\_1.txt} & $x_1, x_2, v_1, v_2 $ & $200$ \\
    Motion-tracked & Coupled oscillator & \texttt{real\_double\_linear\_h\_1.txt} &$x_1,x_2,v_1,v_2$ & $150$ \\
    Simulated & Double pendulum & \texttt{double\_pend\_h\_1.txt}\footnote{First (non-chaotic) configuration only} & $\theta_1, \theta_2, \omega_1, \omega_2 $ & $1355$ \\
    Motion-tracked & Double pendulum & \texttt{real\_double\_pend\_h\_1.txt}$^2$ & $\theta_1, \theta_2, \omega_1, \omega_2 $ & $200$ \\
    \bottomrule
  \end{tabular}
  
  \caption{\label{tab:real-instances}The problem instances for which we used the datasets from \cite{SchmidtSupport2009}. For each system type we use two different datasets, one generated via simulation, the other by motion-tracking the real system.}
\end{table}

\section{Results}
Table \ref{tab:solved} shows the number of successful runs (from 10
independent runs) for each problem instance and algorithm
configuration. A run is considered successful if the IVP solution for
the identified ODE system has an SNMSE $< 0.01$. Some of the instances
can be solved easily with all configurations. Overall the
configuration I$_{\operatorname{opt}}$+D$_{\operatorname{opt}}$ is the
most successful. With this configuration we are able to produce
solutions for all of the 19 instances with a high probability. The
beneficial effect of gradient-based optimization of parameters is
evident from the much larger number of successful runs.

Notably, when we fit the expressions to the approximated derivatives
using partitioning (configurations D and D$_{\operatorname{opt}}$) the
success rate is low. The reason is that IVP solutions might deviate strongly when we
use partitioning to fit of the individual equations to the
approximated derivatives. To achieve a good fit the
causal dependencies must be represented correctly in the ODE
system. This is not enforced when we use partitioning.

\begin{table}
  \centering
  \begin{tabular}{lcccccc}
  Instance & D & I & I+D & D$_{\operatorname{opt}}$ & I$_{\operatorname{opt}}$ & I$_{\operatorname{opt}}$+D$_{\operatorname{opt}}$ \\
  \toprule
ChemicalReaction                         &  7     & 4     & 2     &  9     & 10    & 10    \\
E-CELL                                   &  5     & 0     & 4     &  9     & 10    & 10    \\
S-System                                 &  0     & 0     & 0     &  10    & 10    & 10    \\
Lotka-Volterra (three species)           &  0     & 0     & 0     &  0     & 0     & 8     \\
\midrule
Bacterial Respiration                    &  5     & 3     & 3     &  10    & 10    & 10    \\
Bar Magnets                              &  3     & 4     & 5     &  10    & 10    & 10    \\
Glider                                   &  0     & 0     & 0     &  9     & 10    & 10    \\
Lotka-Volterra                           &  0     & 0     & 0     &  1     & 3     & 10    \\
Predator Prey                            &  0     & 0     & 0     &  3     & 10    & 10    \\
Shear Flow                               &  0     & 0     & 0     &  7     & 10    & 10    \\
Van der Pol Oscillator                   &  1     & 0     & 1     &  6     & 10    & 10    \\
\midrule
Linear Oscillator (motion-tracked)       &  0     & 5     & 9     &  1     & 9     & 10    \\
Linear Oscillator (simulation)           &  0     & 0     & 5     &  0     & 10    & 10    \\
Pendulum (motion-tracked)                &  0     & 0     & 0     &  0     & 10    & 10    \\
Pendulum (simulated)                     &  4     & 9     & 9     &  0     & 10     & 10     \\
Double Oscillator (motion-tracked)       &  0     & 0     & 0     &  0     & 0     & 6     \\
Double Oscillator (simulated)            &  0     & 0     & 0     &  0     & 0     & 10    \\
Double Pendulum (motion-tracked)         &  0     & 0     & 0     &  0     & 0     & 7     \\
Double Pendulum (simulated)              &  0     & 0     & 0     &  10    & 0     & 10    \\
\midrule
Total                                    &  21    & 16    & 29    &  85    & 122   & 171 \\
\bottomrule
  \end{tabular}
  \caption{\label{tab:solved}Number of successful runs. A run is
    successful if the SNMSE for the integrated system is $<
    0.01$. Algorithm configurations: numeric differences (D), numeric
    IVP solution (I), combination of numeric differences and IVP
    solution (I+D). The configurations using the subscript \emph{opt}
    include parameter optimization.}
\end{table}

\section{Discussion}
The results of our experiments are encouraging and indicate that it is
indeed possible to identify ODE models for dynamical systems solely
from data using GP and SR. However, there are several aspects that
have not yet been fully answered in our experiments and encourage
further research.

A fair comparison of algorithm configurations would allow the same
runtime for all cases. We have use a similar amount of function
evaluations but have so far neglected the computational effort that is
required for numerically solving the ODE in each evaluation step.
Noise can have a large effect in the numeric approximation of
derivatives. We have not yet studied the effect of noisy measurements.
Another task for future research is the analysis of forecasting
accuracy of the models. So far we have only measured the performance
on the training set.  Finally, for practical applications, it would be helpful
to extend the method to allow input variables (forced systems) as well
as latent variables.

\subsection*{Acknowledgments}
The authors gratefully acknowledge support by the Austrian Research
Promotion Agency (FFG) within project \#867202, as well as the
Christian Doppler Research Association and the Federal Ministry of
Digital and Economic Affairs within the \emph{Josef Ressel Centre for
  Symbolic Regression}


\begin{thebibliography}{10}
\providecommand{\url}[1]{\texttt{#1}}
\providecommand{\urlprefix}{URL }
\providecommand{\doi}[1]{https://doi.org/#1}

\bibitem{bongard2007}
Bongard, J., Lipson, H.: Automated reverse engineering of nonlinear dynamical
  systems. Proc. of the National Academy of Sciences  \textbf{104}(24),
  9943--9948 (2007)

\bibitem{Chen2018}
Chen, T.Q., Rubanova, Y., Bettencourt, J., Duvenaud, D.K.: Neural ordinary
  differential equations. In: Bengio, S., Wallach, H., Larochelle, H., Grauman,
  K., Cesa-Bianchi, N., Garnett, R. (eds.) Advances in Neural Information
  Processing Systems 31, pp. 6571--6583. Curran Associates, Inc. (2018),
  \url{http://papers.nips.cc/paper/7892-neural-ordinary-differential-equations.pdf}

\bibitem{Gaucel2014}
Gaucel, S., Keijzer, M., Lutton, E., Tonda, A.: Learning dynamical systems
  using standard symbolic regression. In: Genetic Programming. pp. 25--36.
  Springer Berlin Heidelberg, Berlin, Heidelberg (2014)

\bibitem{Gronwall1919}
Gronwall, T.: Note on the derivatives with respect to a parameter of the
  solutions fo a system of differential equations. The Annals of Mathematics
  \textbf{20},  292--296 (1919)

\bibitem{Iba2008}
Iba, H.: Inference of differential equation models by genetic programming.
  Information Sciences  \textbf{178}(23),  4453 -- 4468 (2008).
  \doi{https://doi.org/10.1016/j.ins.2008.07.029}

\bibitem{Isermann2011}
Isermann, R., M{\"u}nchhof, M.: Identification of dynamic systems: an
  introduction with applications. Springer, Berlin, Heidelberg (2011)

\bibitem{kommenda2013}
Kommenda, M., Kronberger, G., Winkler, S., Affenzeller, M., Wagner, S.: Effects
  of constant optimization by nonlinear least squares minimization in symbolic
  regression. In: Proceedings of the 15th annual conference companion on
  genetic and evolutionary computation. pp. 1121--1128. ACM (2013)

\bibitem{rackauckas2018comparison}
Rackauckas, C., Ma, Y., Dixit, V., Guo, X., Innes, M., Revels, J., Nyberg, J.,
  Ivaturi, V.: A comparison of automatic differentiation and continuous
  sensitivity analysis for derivatives of differential equation solutions.
  arXiv preprint arXiv:1812.01892  (2018)

\bibitem{Schmidt2008}
Schmidt, M., Lipson, H.: Data-mining dynamical systems: Automated symbolic
  system identification for exploratory analysis. In: 9th Biennial Conference
  on Engineering Systems Design and Analysis, Volume 2: Automotive Systems;
  Bioengineering and Biomedical Technology; Computational Mechanics; Controls;
  Dynamical Systems. ASME, Haifa, Israel (July 2008)

\bibitem{Schmidt2009}
Schmidt, M., Lipson, H.: Distilling free-form natural laws from experimental
  data. Science  \textbf{324}(5923),  81--85 (2009).
  \doi{10.1126/science.1165893}

\bibitem{SchmidtSupport2009}
Schmidt, M., Lipson, H.: Supporting online material for distilling free-form
  natrual laws from experimental data. Online (April 2009)

\bibitem{Sengupta2014}
Sengupta, B., Friston, K.J., Penny, W.D.: Efficient gradient computation for
  dynamical models. Neuroimage  \textbf{98},  521--527 (2014)

\bibitem{topchy2001}
Topchy, A., Punch, W.F.: Faster genetic programming based on local gradient
  search of numeric leaf values. In: Proceedings of the 3rd Annual Conference
  on Genetic and Evolutionary Computation. pp. 155--162. Morgan Kaufmann
  Publishers Inc. (2001)

\bibitem{worm2014}
Worm, T., Chiu, K.: Scaling up prioritized grammar enumeration for scientific
  discovery in the cloud. In: 2014 IEEE International Conference on Big Data.
  pp. 621--626. IEEE (2014)

\end{thebibliography}

\end{document}